\documentclass[11pt]{article}
\usepackage[letterpaper,margin=1in]{geometry}
\usepackage{fontspec}              
\usepackage{amsmath,amssymb}
\usepackage{booktabs}
\usepackage{graphicx}
\usepackage{enumitem}
\usepackage{tikz}
\usetikzlibrary{arrows.meta,positioning}
\usepackage[round,authoryear]{natbib}
\usepackage[colorlinks=true,allcolors=blue]{hyperref}

\newcommand{\vag}{V\=agdhenu}      
\title{\textbf{V\=agdhenu: A V\d{r}tta (Meter) Aware \'Sloka-to-Chant (TTS) System for Sanskrit}}

\author{Prathosh A P\\[2pt]
  Indian Institute of Science, Bengaluru\\
  \texttt{prathosh@iisc.ac.in}}
\date{Preprint. \today}

\begin{document}
\maketitle

\begin{abstract}
We present \vag{}, a v\d{r}tta (meter) aware \'sloka-to-chant system for Sanskrit, that is, a text-to-speech system that maps a metrical verse to its chanted par\=aya\d{n}a recitation at high fidelity, together with the two large deployments that exercised it end to end. The system is an experience report rather than a new architecture: it takes an off-the-shelf flow-matching text-to-speech backbone \citep{chen2024f5,ai4bharat_indicf5} and a large-scale neural vocoder \citep{lee2023bigvgan}, and adds the components that a faithful Sanskrit chant pipeline actually needs. These are a script-aware frontend that routes Sanskrit through Kannada orthography to avoid the Hindi-style schwa deletion that Devanagari triggers in Indic models; a frontend that obeys subtle Sanskrit phonological rules, namely visarga sandhi including the jihv\=am\=ul\=iya and upadhm\=an\=iya allophones, the aspiration contrast of alpapr\=a\d{n}a and mah\=apr\=a\d{n}a, the three sibilants dantya s, m\=urdhanya \d{s}, and t\=alavya \'s kept distinct, and homorganic (parasavar\d{n}a) realization of anusv\=ara; a v\d{r}tta-aware mechanism that detects the meter and selects an exactly matched reference under what we call the half-reference rule; and a quality-control stack built around source validation rather than forced alignment. We report a clean negative result that shaped the whole system: in a self-infilling flow-matching backbone, a text-side prosody conditioner is architecturally inert, because the model recovers pitch from the context mel and the conditioning embedding receives no gradient. The reference clip and a voice-steering retrain are the only working prosody levers. We also report a comparative lineage across four architecture families (StyleTTS2, VITS2, Matcha-TTS, and the flow-matching backbone), where each earlier family hit a ceiling on conjunct rendering or prosody that a five hour clone on the flow-matching backbone cleared at an expert mean opinion score of about 4.6. The system shipped two deployments: a 32 chapter, 5{,}183 verse video corpus of the Mah\=abh\=arata T\=atparya Nir\d{n}aya (about 17.5 hours), and an audio mobile application covering the \'Sr\=imad Bh\=agavatam (about 18{,}000 verses across 12 books). We release the frontend, inference and training code, model weights, a single-speaker chant dataset, and an interactive demonstration. Code, weights, data, and demo are public.\footnote{Code: \url{https://github.com/prathoshap/vagdhenu}. Weights and dataset: \url{https://huggingface.co/prathoshap}. Demo: \url{https://huggingface.co/spaces/prathoshap/vagdhenu-demo}.}
\end{abstract}

\section{Introduction}
Classical Sanskrit recitation, par\=aya\d{n}a, is a chanted rather than a read register. A faithful synthesizer must hold long vowels, sustain a terminal visarga, articulate retroflex and aspirated consonants, render dense consonant conjuncts cleanly, and respect the metrical structure of the verse. None of these is well served by general-purpose text-to-speech, and there is essentially no chant-domain training data available off the shelf. The problem is therefore doubly hard: it is low-resource, and the target prosody is a specialized melodic contour rather than ordinary read speech.

This report describes a system, \vag{}, that solves the practical version of this problem well enough to ship two large deployments, and it documents the design decisions, the dead ends, and the one negative result that turned out to be the most useful finding. We do not claim a new model. We claim an honest account of what it takes to build a faithful Sanskrit chant pipeline on top of current open backbones, what works, and what is architecturally out of reach.

Our framing is that of an experience report. The evidence we offer is the comparative lineage across architecture families, a reproducible production system, two shipped artifacts at real scale, and a public release of code, weights, data, and a live demonstration. Formal listening studies are limited to expert evaluation, which we state plainly and treat as a limitation rather than a result.

\subsection{Contributions}
\begin{itemize}[leftmargin=1.4em,itemsep=2pt]
  \item \textbf{A v\d{r}tta-aware \'sloka-to-chant system.} The system detects the meter of each verse and uses it to govern synthesis, namely to select an exactly matched reference under the half-reference rule, to set the per-meter duration budget, and to place the caesura (yati). This is a reference-side conditioning path that works, as opposed to a learned text-side meter conditioner, which we show is inert.
  \item \textbf{Faithful rendering of subtle Sanskrit phonology and sandhi.} The system renders distinctions that general text-to-speech ignores: visarga sandhi (utva, rutva, lopa, and satva) together with the jihv\=am\=ul\=iya and upadhm\=an\=iya allophones of the visarga before velars and labials; the aspiration contrast between alpapr\=a\d{n}a and mah\=apr\=a\d{n}a (unaspirated against aspirated stops); the three sibilants, dantya s, m\=urdhanya \d{s}, and t\=alavya \'s, kept distinct; the full retroflex series; homorganic (parasavar\d{n}a) realization of anusv\=ara; and vocalic \d{r}. Dense conjuncts, including retroflex aspirates, render correctly, which is the class the earlier architecture families could not crack. The frontend that encodes these rules is released as open source.
  \item \textbf{A comparative architecture lineage on a hard domain.} We document four architecture families evaluated on the same data, the conjunct or prosody ceiling each one hit, and why a five hour clone on a flow-matching backbone cleared them. The lesson is that, for this domain, the backbone, not the data, was the bottleneck.
  \item \textbf{A negative result on prosody controllability.} We show, through several controlled probes, that a text-side prosody conditioner is architecturally inert in a self-infilling flow-matching backbone, and we explain why. We identify the reference clip and a voice-steering retrain as the only working levers, and formalize the half-reference rule for reference construction.
  \item \textbf{A quality-control finding.} We report that forced-alignment quality control is low-signal at production scale, while the genuine defects came almost entirely from the source text. We describe the lightweight gates that replaced it.
  \item \textbf{Two deployments and a public release.} We describe a 5{,}183 verse video corpus and an 18{,}000 verse audio application, and we release the frontend, inference and training code, model weights, a single-speaker chant dataset, and an interactive demonstration.
\end{itemize}

\section{Related work}
\paragraph{Neural text-to-speech and vocoders.}
Modern neural speech synthesis moved from autoregressive acoustic models and neural vocoders \citep{oord2016wavenet,shen2018tacotron2} to non-autoregressive and end-to-end designs. End-to-end variational models \citep{kim2021vits,kong2023vits2} and style-diffusion models \citep{li2023styletts2} reach near human quality on read speech. Vocoding is now dominated by adversarial mel-to-waveform models \citep{kong2020hifigan,lee2023bigvgan} and Fourier-domain alternatives \citep{siuzdak2024vocos}. We use these as building blocks rather than novelty, and we report below a concrete case where the choice of vocoder is not interchangeable for our domain.

\paragraph{Flow-matching synthesis.}
Conditional flow matching \citep{lipman2023flow,tong2024otcfm} provides a stable, non-autoregressive generative objective that has been applied to speech with strong results, including a fast architecture with explicit duration \citep{mehta2024matcha} and a diffusion-transformer backbone trained for in-context cloning \citep{chen2024f5}. The transformer backbone \citep{peebles2023dit} that \citet{chen2024f5} adopt performs mel infilling and clones a reference clip in context. Our production model is a multilingual Indic instance of this backbone \citep{ai4bharat_indicf5}. A property of this family, which we make central, is that it has no native duration or pitch head and recovers prosody from the context mel during infilling.

\paragraph{Indic and low-resource synthesis.}
Indic text-to-speech has advanced through multilingual data collection and shared backbones \citep{ai4bharat_indicf5,ai4bharat_indicparler}. Sanskrit specifically is low-resource and orthographically demanding: dense conjuncts, retroflex and aspirated contrasts, visarga, anusv\=ara, and vocalic vowels all matter for correctness. We address these with a dedicated frontend rather than with more data.

\paragraph{Controllable and expressive prosody.}
Text-conditioned and description-conditioned prosody control has been explored through phoneme-level language models \citep{li2023plbert}, supervised semantic tokens \citep{du2024cosyvoice}, and natural-language style prompts \citep{lyth2024parler}. We probed two of these directions for chant control and report why they did not fit our constraints, namely overfitting on a small chant corpus and content drift under description-based steering. Our negative result on text-side conditioning is specific to the self-infilling flow-matching backbone and complements this literature.

\paragraph{Alignment and computational Sanskrit.}
Forced alignment \citep{mcauliffe2017mfa} and Conformer-based recognition \citep{gulati2020conformer,ai4bharat_indicconformer} are standard tools for data preparation and quality control. We use alignment for duration extraction but report that, as a synthesis quality gate at scale, it was low-signal. Computational treatment of Sanskrit, including segmentation and sandhi analysis, is an established area \citep{goyal2016heritage}; our frontend implements a focused, synthesis-oriented subset.

\section{The Sanskrit text frontend}
The frontend converts sa\d{m}hit\=a (already sandhified, continuous) Sanskrit text into the exact token stream the backbone synthesizes, and into a transliteration suitable for alignment. It is small, deterministic, and released as open source.

\subsection{Script routing}\label{sec:routing}
The single most important rule is orthographic, not acoustic. Indic backbones trained across Indian scripts treat raw Devanagari with Hindi phonotactics and delete the inherent schwa, which is wrong for Sanskrit. Rendering the same text in \emph{Kannada} script avoids this, because Kannada orthography is read with the inherent vowel intact. We therefore transliterate all input to Kannada before synthesis. The mapping is lossless and round-trips, so display text and manifests retain the original Devanagari while the model receives Kannada. As a practical bonus, accepting any Brahmic script as input (Devanagari, Bengali, Gurmukhi, Gujarati, Oriya, Tamil, Telugu, Kannada, Malayalam, Grantha) becomes a single transliteration step.

\subsection{Phonological and sandhi fidelity}
On top of routing, the frontend obeys a set of Sanskrit phonological rules that general text-to-speech ignores but that a faithful recitation requires. We list them because together they are a distinguishing capability of the system.

\paragraph{Visarga sandhi and its allophones.} Internal visarga sandhi (utva, rutva, lopa, and satva in the cases where it is phonetically real) is applied by default and can be disabled for already-sandhified citation (pram\=a\d{n}a) text. A da\d{n}\d{d}a-final visarga is realized by the length of the preceding vowel: after a short vowel it is echoed as a light aspirate, while after a long vowel it is left bare so that it elongates naturally. The visarga allophones jihv\=am\=ul\=iya (before velars k and kh) and upadhm\=an\=iya (before labials p and ph) are preserved, the backbone having learned them acoustically rather than through a substitution.

\paragraph{Aspiration and the sibilants.} The system keeps the aspiration contrast between alpapr\=a\d{n}a and mah\=apr\=a\d{n}a stops (for example k against kh, t against th, p against ph) audible, which the recording protocol of Section~\ref{sec:data} reinforces. It also keeps the three sibilants distinct: dantya (dental) s, m\=urdhanya (retroflex) \d{s}, and t\=alavya (palatal) \'s. The retroflex series \d{t}, \d{t}h, \d{d}, \d{d}h, \d{n} is the hardest class for a synthesizer and is the one the production backbone renders correctly where the earlier families did not.

\paragraph{Anusv\=ara and vocalic vowels.} Anusv\=ara is realized as the homorganic (parasavar\d{n}a) nasal of the following consonant, looking past intervening spaces. Long vocalic-r is handled explicitly to avoid a mispronunciation in the target script, and a focused rule treats vocalic r before a conjunct. Editorial parentheticals, which are not recited, are stripped before metrical analysis so they do not corrupt meter detection. The full rule list is in Appendix~B.

\subsection{Meter and syllable weight}
Each verse is scanned into a laghu and guru (light and heavy syllable) sequence, and matched to a meter (v\d{r}tta) signature. Meter signatures are self-calibrated from the reference bank, matching is done per quarter-verse (p\=ada) with quarter-final anceps ignored, and mixed-meter verses are handled per half-verse. Meter drives reference selection (Section~\ref{sec:refmech}) and the per-syllable duration budget.

\section{Model lineage: four eras}\label{sec:lineage}
The system passed through four architecture families on the same task and data before settling. We summarize them because the comparison is itself a finding: each earlier family hit a hard ceiling on conjunct rendering or prosody, and the flow-matching backbone cleared both with far less data. Table~\ref{tab:lineage} gives the summary. Only the fourth family is in production; the first three are retired and are documented so the dead ends are not re-walked. The detailed experiment ledger is in Appendix~C.

\begin{table}[t]
\centering
\small
\begin{tabular}{@{}clccp{4.7cm}@{}}
\toprule
Era & Backbone & Params & MOS & Ceiling reached \\
\midrule
1 & StyleTTS2 & 70M & 3.0 to 4.2 & conjuncts muffled; English priors loaded silently \\
2 & VITS2 & 39.9M & $>$ Era 1 & conjuncts still muffled, traced to data sparsity \\
3 & Matcha-TTS & 18.2M & 4.2 to 4.3 & conjuncts improved; meter prosody still missing \\
4 & Flow-matching (IndicF5) & 337M & 4.6 & none reached; the production lineage \\
\bottomrule
\end{tabular}
\caption{Architecture lineage on the same single-speaker Sanskrit data. MOS values are expert single-listener estimates and are not directly comparable across rows; they indicate the trajectory, not a controlled benchmark. Era 4 cleared the conjunct class, including retroflex aspirates, that the earlier families could not, with a five hour clone rather than more data.}
\label{tab:lineage}
\end{table}

\paragraph{Era 1: StyleTTS2.} A fork of \citet{li2023styletts2} with a Sanskrit phoneme-level encoder \citep{li2023plbert} and forced-alignment durations reached a usable quality, but conjuncts stayed muffled and the early pilot suffered from silently loaded English priors. The diagnosis that conjunct elongation is a decoder limit rather than an alignment problem carried forward.

\paragraph{Era 2: VITS2.} An end-to-end model \citep{kong2023vits2} trained from scratch produced crisper output but the same conjunct ceiling, which we traced to data sparsity for specific retroflex consonants rather than to architecture.

\paragraph{Era 3: Matcha-TTS.} A flow-matching acoustic model \citep{mehta2024matcha} with a large-scale vocoder \citep{lee2023bigvgan} improved conjuncts materially. The vocoder bake-off here is where we first established that a particular vocoder choice removed a hum and smoothed the repha, a finding that recurs in production.

\paragraph{Era 4: flow-matching backbone.} A diffusion-transformer flow-matching backbone trained for in-context cloning \citep{chen2024f5}, in its multilingual Indic instance \citep{ai4bharat_indicf5}, cloned Sanskrit chant zero-shot, and a five hour fine-tune cleared the full conjunct class at an expert score of about 4.6. Two implementation facts were load-bearing: a normalization bug in the installed backbone had to be patched for the text encoder to function, and Devanagari input triggered schwa deletion, which the Kannada routing of Section~\ref{sec:routing} fixes.

\section{The production system}

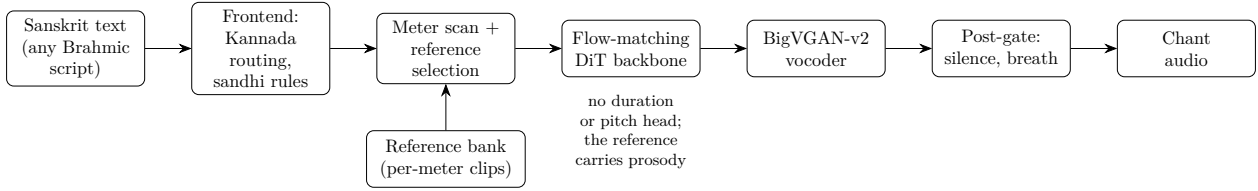
\begin{figure}[t]
\centering
\resizebox{\columnwidth}{!}{%
\begin{tikzpicture}[
  node distance=8mm and 9mm,
  box/.style={draw, rounded corners, align=center, font=\small, minimum height=11mm, inner sep=4pt, text width=24mm},
  ann/.style={font=\footnotesize, align=center},
  >={Stealth[length=2.4mm]}]
\node[box] (in) {Sanskrit text\\(any Brahmic script)};
\node[box, right=of in] (fe) {Frontend:\\Kannada routing,\\sandhi rules};
\node[box, right=of fe] (ref) {Meter scan +\\reference\\selection};
\node[box, right=of ref] (dit) {Flow-matching\\DiT backbone};
\node[box, right=of dit] (voc) {BigVGAN-v2\\vocoder};
\node[box, right=of voc] (post) {Post-gate:\\silence, breath};
\node[box, right=of post] (out) {Chant\\audio};
\draw[->] (in)--(fe);   \draw[->] (fe)--(ref);  \draw[->] (ref)--(dit);
\draw[->] (dit)--(voc); \draw[->] (voc)--(post); \draw[->] (post)--(out);
\node[box, below=9mm of ref, text width=28mm] (bank) {Reference bank\\(per-meter clips)};
\draw[->] (bank)--(ref);
\node[ann, below=2mm of dit, text width=34mm] {no duration or pitch head;\\the reference carries prosody};
\end{tikzpicture}}
\caption{The \vag{} inference pipeline. Sanskrit text in any Brahmic script is transliterated to Kannada and normalized by the frontend, the meter is scanned and an exactly matched reference is selected from the bank, the flow-matching backbone synthesizes a mel by in-context infilling against the reference, the vocoder produces the waveform, and a post-gate handles silence and breath. Because the backbone has no duration or pitch head, the reference clip carries voice, melodic contour, and pace.}
\label{fig:pipeline}
\end{figure}

\subsection{Backbone and vocoder}
The production model is the flow-matching backbone \citep{chen2024f5,ai4bharat_indicf5}, a diffusion transformer of about 337M parameters with no native duration or pitch head, paired with a large-scale neural vocoder \citep{lee2023bigvgan}. The vocoder is mandatory rather than interchangeable: a Fourier-domain alternative \citep{siuzdak2024vocos} introduced a long-vowel phase artifact that the adversarial vocoder does not, which we verified directly. We also found that residual wobble and jitter live in the acoustic model, not in the vocoder, which localizes where each class of fix has to be applied.

\subsection{The reference mechanism}\label{sec:refmech}
Because the backbone has no duration or pitch head, the reference clip carries voice, melodic contour, and pace. A per-clip duration budget sets only the total mel length, and the backbone pads rather than stretches, so an over-budget request parks silence inside the verse. Reference selection is therefore central. We select a reference by exact per-quarter-verse laghu and guru match, cleanest harmonics-to-noise ratio, and a clean da\d{n}\d{d}a ending. Because that match is keyed to the verse's meter, reference selection is the system's v\d{r}tta-aware conditioning path: the detected meter governs which prosodic template the backbone clones, and it also sets the per-meter duration budget and the caesura position. This is conditioning on the meter through the reference, as opposed to a learned text-side meter embedding, which Section~\ref{sec:prosody} shows is inert.

\paragraph{The half-reference rule.}
The reference text must match the spoken span of the reference audio on a clean word or da\d{n}\d{d}a boundary. A clean half-hemistich of about seven seconds outperforms a full fifteen second \'sloka, because a full verse ending on a cadence causes the first synthesized syllable to be swallowed. A half-\'sloka reference repeated three times, to about twenty seconds, markedly improves the melodic contour. This rule, together with whole-verse rendering for short meters, is the practical lever for prosody quality.

\section{Training and compute}
\label{sec:training}
\paragraph{Backbone.} The production backbone is a diffusion-transformer flow-matching model with dimension 1024, depth 22, 16 attention heads, a feed-forward multiplier of 2, a text dimension of 512, and a convolutional kernel of 4, for about 337M parameters \citep{chen2024f5,ai4bharat_indicf5}. The objective is the conditional-flow-matching velocity regression on a masked mel-infilling task, with classifier-free guidance trained by dropping the audio condition with probability 0.3 and the text condition with probability 0.2.

\paragraph{Voice fine-tune.} The single-speaker voice is produced by fine-tuning the multilingual backbone on the chant corpus, warm-started from the Indic checkpoint, at a learning rate of $1\mathrm{e}{-5}$, a frame batch of about 12800 to 19200 (roughly 28GB of memory), in bfloat16, for several hundred epochs, with checkpoints saved every 500 to 2000 steps. Training is distributed-data-parallel across two GPUs, and a 100-band mel at 24~kHz is used throughout. A short voice-steering retrain on 179 paired clips (Section~\ref{sec:prosody}) then makes the voice more responsive to the reference; this is the shipped production voice, with the reference-driven fine-tune kept as a fallback.

\paragraph{Vocoder fine-tune.} The vocoder is the large-scale adversarial vocoder \citep{lee2023bigvgan} fine-tuned on the backbone's own mel representation, then averaged into an exponential-moving-average soup of the last several checkpoints. It must be loaded into the correct generator class, because a mismatched class silently half-loads the weights and yields monotone output.

\paragraph{Compute and serving.} Training and rendering ran on hosts with two RTX A6000 (48GB) GPUs each, with a second such host used for a four-GPU render fan-out on the large deployments. The deployed model is about 449M parameters in total (backbone plus vocoder), roughly 900MB in half precision, with a peak inference memory of about 2.5GB. At 64 function evaluations the real-time factor is about 1.24, and at 32 it is about 0.63, which is the serving setting; a GPU is required.

\section{Prosody controllability: a negative result}\label{sec:prosody}
The most useful finding of the project is a negative one. We wanted designable, verse-independent prosody, namely the ability to specify a melodic contour from the text side. In the self-infilling flow-matching backbone this is architecturally unreachable, and we established it through several controlled probes.

The mechanism is as follows. During infilling, the backbone is conditioned on the clip's own context mel, which already reveals the pitch. A text-side conditioning embedding, whether it encodes the gana (laghu and guru pattern) or a swara (melodic) template, is therefore redundant, and it receives essentially no gradient. In our probes the conditioning gradient norm stayed near 0.02, obedience with the conditioner on was indistinguishable from off, and the effect at high classifier-free guidance was about zero semitones. The conclusion is to never re-attempt text-side embedding tuning on this backbone. We state the controllability picture as three facts:

\begin{itemize}[leftmargin=1.4em,itemsep=2pt]
  \item Contour is text-driven. Same-reference, different-verse pitch correlation is about 0.13, while different-reference, same-verse correlation is about 0.40 to 0.54.
  \item The working levers are the reference clip (supply the desired prosody as a clean, exactly matched reference) and a voice-steering retrain that makes the voice more reference-responsive.
  \item Classifier-free guidance trades expression against tremor, because both ride the same stochastic pitch and jitter. A low value is expressive but tremulous; a high value is clean but flatter.
\end{itemize}

We also probed two external controllable lines and dropped both: a semantic-token zero-shot synthesizer \citep{du2024cosyvoice} overfit on a small chant corpus, and a description-conditioned model \citep{lyth2024parler,ai4bharat_indicparler} introduced content drift that the flow-matching backbone does not. Arbitrary designed prosody would require a different architecture, namely an explicit duration predictor and length regulator, which we designed but did not build.

\section{Data}\label{sec:data}
A central design question for a chant corpus is which verses to record. Natural Sanskrit corpora are dominated by a single meter, so a corpus drawn by sampling text at random would teach the model one meter well and the rest poorly, and would under-cover the phonetically hard sounds. We therefore designed the corpus for metrical and phonetic balance.

\subsection{Source texts}
The verses were drawn from classical Sanskrit metrical works chosen for metrical diversity, principally the Mah\=abh\=arata T\=atparya Nir\d{n}aya (about 5{,}100 verses), the Sumadhvavijaya (16 cantos, about 1{,}000 verses, which adds m\=alin\=i and atijagat\=i), and a purpose-built meter-diverse recording sheet of 62 verses across 10 meters, supplemented by a sragdhar\=a source for that meter. From these pools we built a recording manifest of about 12{,}487 half-verse segments (roughly 40 hours), each annotated with its Devanagari and transliterated text, its per-syllable laghu and guru feature, its meter, a reference, and an estimated duration. A few sources were excluded when their digital text was corrupt, for example a sragdhar\=a stotra with a systematic repha transposition.

\subsection{V\d{r}tta-balanced corpus design}
The corpus was designed to flatten the meter distribution and to meet explicit coverage thresholds for the sounds that are hardest to synthesize. The targets were at least 300 occurrences of each retroflex consonant, at least 500 long vowels, at least 200 repha and r-clusters, at least 150 of each hard conjunct (for example k\d{s}a, j\~na, tra, dra, dhra, hma, hna), and at least 100 occurrences of every one of the 57 phonemes. Two meters, rucir\=a and m\=alin\=i, were deliberately held out so that synthesis on an unseen meter could be tested. A corpus-coverage audit tool checked these thresholds against the candidate manifest before recording. The original target was 25 hours of clean audio, but a five hour recording already reached the quality ceiling (Section~\ref{sec:lineage}), so the full target was deprioritized rather than completed.

\subsection{Recording protocol}
The production voice is a single new speaker recorded chant-first, with no warm-start, in a controlled room with a fixed microphone position, a quiet noise floor, and lossless capture at 24-bit, 48~kHz, downsampled to 24~kHz for training. A tonic drone is monitored during recording to anchor the tonic across sessions but is not captured. The protocol encodes several hard-won rules. There are no audible gaps between words within a quarter-verse, so each quarter is one continuous breath group, and a pause occurs only at a da\d{n}\d{d}a. Long meters take one silent breath at the caesura (yati), at the same fixed position in every clip. Long vowels are held, terminal visarga is sustained, retroflex and aspirated consonants are articulated, and geminates are held. Sessions interleave several meters to keep the prosody manifold broad. A recurring lesson is that audible word gaps in an early pilot taught the model to read a space as a pause, which the protocol then corrected.

\subsection{The released dataset}
The single-speaker chant dataset we release contains about 1{,}467 clips, roughly 5.3 hours, in two recording styles (a measured style and a second, differently paced style), each clip paired with its text in a metadata file. The kept portion of the primary style is about 3.0 hours over 764 verses and about 33{,}700 syllables, spanning roughly nine meter families (Table~\ref{tab:corpus}). The dataset is released under a Creative Commons attribution license; it is the author's own voice and verses, so there is no third-party consent question.

\begin{table}[t]
\centering
\small
\begin{tabular}{@{}lr@{}}
\toprule
Meter family & Share of the primary-style corpus \\
\midrule
anu\d{s}\d{t}ubh & 31\% \\
tri\d{s}\d{t}ubh & 17\% \\
jagat\=i / va\d{m}\'sastha & 13\% \\
sragdhar\=a & 11\% \\
vasantatilak\=a & 9\% \\
m\=alin\=i & 6\% \\
rucir\=a & 3\% \\
\'s\=ard\=ulavikr\=i\d{d}ita & 2\% \\
17-syllable meters & 2\% \\
\bottomrule
\end{tabular}
\caption{Meter distribution of the primary-style recording corpus (about 3.0 hours, 764 verses). The distribution is deliberately flattened relative to natural text, where anu\d{s}\d{t}ubh alone is 48 to 87 percent, so that long meters are represented. Rucir\=a and m\=alin\=i double as held-out generalization meters.}
\label{tab:corpus}
\end{table}

\section{Production pipeline and quality control}\label{sec:qc}
The render pipeline loads the backbone and vocoder once and processes per-hemistich clips, with a multi-GPU fan-out for marathon runs. A post-render gate removes the vocoder onset transient and tail ring, and is fricative-aware and stop-aware so that it does not clip a leading fricative or a final stop burst.

The quality-control finding is the operationally important one. We built a forced-alignment quality gate and ran it on the full video corpus of about 10{,}220 clips. It produced about 1{,}648 flags, of which a manual sample found essentially none to be real defects. The alignment gate's false-positive floor, driven by conjuncts, fricative onsets, long meters, and reference priming, dominated. Only a cheap duration gate caught the genuine collapses. The conclusion, which we carried into the second deployment, is to drop forced-alignment quality control and keep a duration gate, an internal-silence gate for spurious pauses, a small best-of-N reseed of the flagged minority, and recognition-based error detection, together with structural and source validation. The dominant source of real defects was the source text itself, namely numbering duplicates and gaps, merged entries, editorial brackets, and transcription typographical errors, all of which the synthesizer reproduces faithfully.

\section{Case study I: a video corpus}
The first deployment is the Mah\=abh\=arata T\=atparya Nir\d{n}aya, a 32 chapter work of 5{,}183 verses, rendered as 32 videos with bilingual cards (Devanagari and Kannada), per-hemistich highlighting, and a tonic drone, for a total of about 17.5 hours. The marathon render across four GPUs took about 13 hours and produced 32 of 32 chapters cleanly, with asynchronous review and surgical single-verse re-renders for the few defects. The defects that occurred were almost all source-numbering and structural issues, consistent with Section~\ref{sec:qc}. A practical lesson from publishing is that a published video file cannot be replaced in place, so issues must be caught before publication, which motivates an unlisted soft launch.

\section{Case study II: an audio application}
The second deployment targets the \'Sr\=imad Bh\=agavatam, about 18{,}000 verses across 12 books and roughly 345 chapters, as an audio-first mobile application rather than video. Two design changes followed from the first deployment. We dropped the video half entirely, and we replaced forced-alignment quality control with the structural and source-validation stack of Section~\ref{sec:qc}. A structural audit ran before any rendering and confirmed that the source was exceptionally clean, with no empty text blocks and no chapter, verse, or book numbering mismatches against the embedded colophons, leaving only a small number of benign cosmetic issues. A meter census over the 16{,}017 rendering units (Table~\ref{tab:meter}) drove reference selection: the corpus is dominated by anu\d{s}\d{t}ubh, with a long tail of named meters and about 1{,}348 mixed-meter verses handled per half-verse.

\begin{table}[t]
\centering
\small
\begin{tabular}{@{}lr@{}}
\toprule
Class or meter & Units \\
\midrule
Metrical verse & 14{,}042 \\
Prose (gadya) & 559 \\
Short connectives & 1{,}416 \\
\midrule
anu\d{s}\d{t}ubh & 11{,}464 \\
vasantatilak\=a & 649 \\
jagat\=i-upaj\=ati & 382 \\
va\d{m}\'sastha & 371 \\
upaj\=ati & 337 \\
mixed-meter (per half-verse) & 1{,}348 \\
\bottomrule
\end{tabular}
\caption{Meter census of the second deployment (\'Sr\=imad Bh\=agavatam), 16{,}017 rendering units across 12 books and about 345 chapters. The top block is the coarse class split; the lower block lists the most frequent meters. Roughly 185 ardhasama verses lacked a matching reference in the bank and used a nearest-length fallback.}
\label{tab:meter}
\end{table}

The hardest sub-problem was prose (gadya), which the production model was not built for. A chunking rule, locked by ear, strips colophons and parentheticals, breaks only on safe word boundaries (never inside a word), and packs whole words to a target length, with care taken at de-sandhied junctions where a space is not a safe break. The audio application ships per-verse audio with optional drone mixing, a compact on-device database, and a per-chapter karaoke timing track so that the recited line is highlighted as it plays. The application is the audio companion that the project had pointed at from the start.

\section{Evaluation}\label{sec:eval}
We are explicit that formal evaluation is limited, and we treat this as a limitation rather than a result. The headline quality figure is an expert single-listener mean opinion score of about 4.6 for the production clone, with all consonant conjuncts, including retroflex aspirates, rendered correctly, which is the class the earlier architecture families could not crack. For reference, expert estimates placed strong commercial systems in the 4.0 to 4.6 range and human recitation at about 4.6 to 4.7 on the same informal scale. The architecture lineage of Table~\ref{tab:lineage} is the comparative evidence, run on the same data across four families. The prosody claims of Section~\ref{sec:prosody} are supported by the pitch-correlation figures reported there. We provide curated audio samples with the public release so that readers can assess quality directly, and we leave a controlled multi-listener study, with blind baselines and confidence intervals, as future work. The objective measures most relevant to this domain, namely a conjunct-accuracy rate, a recognition-based character error rate, duration fidelity, and the pitch-correlation table, are the natural next additions.

\section{Open release: \vag{}}
We release the system publicly under the name \vag{} (v\=agdhenu, the wish-granting cow of speech). The release is scoped to the text-to-speech system, namely the frontend, the inference path, and the training code, and excludes the corpus-rendering, karaoke, and video-assembly pipelines of the two deployments. It is a single-speaker release built entirely from the author's own voice and data, so there is no third-party consent question; a pilot reciter from an earlier era is anonymized and excluded.

The components are: a code repository with the frontend, a configurable inference path, the reference bank, and training scripts, under a permissive license; a model repository with the production voice, a reference-driven fallback voice, and the fine-tuned vocoder, redistributable under the permissive licenses of the base models; a single-speaker chant dataset of about 1{,}467 clips, roughly 5.3 hours across two recording styles, under a Creative Commons attribution license; and an interactive demonstration that accepts any Brahmic script, auto-detects the meter, falls back to a default meter when detection is unsure, and offers a menu of sample verses. The demonstration runs on on-demand GPU hosting. All four components are public.

\section{Limitations and ethics}
The evaluation is the main limitation, as stated in Section~\ref{sec:eval}: the quality figures are expert estimates, not a controlled multi-listener study, and the lineage MOS values are not directly comparable across rows. Several open problems remain. Designable, verse-independent prosody is out of reach on the current backbone and would require a different architecture. Voice-agnostic chant, namely zero-shot chant in an arbitrary voice, would require a multi-reciter chant fine-tune. Repeated-syllable depth beyond about four is unrecoverable even with priming, and a few rare artifacts, including a long-vowel bare-visarga edge case, are handled per clip rather than globally.

On ethics and intended use, the production voice is the author's own, recorded with consent, and the released dataset is the author's own recordings. The system synthesizes a specific liturgical register; it is intended for recitation content, study recordings, and accessibility, and it is not a general voice-cloning tool. The released weights are single-speaker. Source-text editions are cited in the deployments, and editorial issues in the source are documented rather than silently corrected.

\section{Conclusion}
We presented \vag{}, a v\d{r}tta-aware \'sloka-to-chant system for Sanskrit, and an honest account of building it. The contributions are a frontend that obeys subtle Sanskrit phonology, namely visarga sandhi with its jihv\=am\=ul\=iya and upadhm\=an\=iya allophones, the aspiration contrast of alpapr\=a\d{n}a and mah\=apr\=a\d{n}a, and the three distinct sibilants, while routing the language through Kannada orthography; a v\d{r}tta-aware mechanism that conditions synthesis on the detected meter through reference selection rather than a learned embedding; a comparative architecture lineage showing that the backbone rather than the data was the bottleneck for this domain; a clean negative result that a text-side prosody conditioner is architecturally inert in a self-infilling flow-matching backbone; a quality-control finding that source validation matters more than forced alignment at scale; and two shipped deployments at real scale. We release the system, its weights, a chant dataset, and a live demonstration so that the result is reproducible and usable. The open problem we most want to see addressed is designable prosody, which on current backbones requires an architecture with an explicit duration and length model.

\section*{Acknowledgments}
We thank the custodians of the source-text editions used in the two deployments. A pilot reciter from an earlier phase of the project is gratefully acknowledged and remains anonymized at their preference.

\bibliographystyle{plainnat}
\bibliography{references}

\appendix
\section{Locked inference parameters}
The production configuration, as built, uses the flow-matching backbone with the Euler solver, a number-of-function-evaluations setting of 64, a sway-sampling coefficient of about $-0.7$, a speed factor of 0.90, the large-scale vocoder, and a per-clip duration budget equal to the reference length plus the syllable count times a per-meter seconds-per-syllable value taken from the reference bank. Classifier-free guidance is set high (about 3.0) for the batch renderer to suppress tremor, and lower for single-verse work; this is the expression-versus-tremor trade of Section~\ref{sec:prosody}. Post-processing applies a head and tail gate, silence compression, and a da\d{n}\d{d}a breath, with an optional tonic drone for the video deployment.

\section{Frontend rules}
The frontend applies, in order: transliteration of any Brahmic input to a working script; punctuation stripping; conversion to a transliteration scheme; optional internal sandhi (utva, rutva, lopa, and satva where phonetically real); homorganic realization of anusv\=ara as the nasal of the following consonant; da\d{n}\d{d}a-final visarga handling (echoed after a short vowel, bare after a long vowel); explicit handling of long vocalic-r; a metathesis rule for two awkward conjuncts that the backbone renders poorly; and stripping of editorial parentheticals before metrical analysis. Citation (pram\=a\d{n}a) verses are rendered with sandhi; already-sandhified text is rendered without it.

\section{Experiment ledger (E0 to E80h)}
The system was developed through about eighty numbered experiments across the four eras of Section~\ref{sec:lineage}. The ledger below is the distilled record. Infrastructure detail, run identifiers, and the pilot reciter's identity are omitted; the pilot reciter is the source of the Era 1 to Era 3 single-speaker data and is anonymized.

\paragraph{Era 1, StyleTTS2 (E0 to E8).}
\textbf{E0} StyleTTS2 pilot with an English warm-start, MOS about 3.0; English language priors were loaded silently, and the anu\d{s}\d{t}ubh-heavy data generalized poorly. \textbf{E1} v0\_mfa: a Sanskrit phoneme-level encoder, forced-alignment-direct durations, and a fine-tune from the pilot reached MOS 4.2 or above, with the duration loss roughly halved. \textbf{E2} an inference pad-symmetry fix gave no audible change. \textbf{E3} cluster-duration scaling was worse, which showed that conjunct elongation is a decoder limit rather than alignment over-allocation. \textbf{E4} a forced-alignment self-audit confirmed that the chanter genuinely holds conjuncts long. \textbf{E5} a decoder-only fine-tune gave no gain. \textbf{E6} a Conformer aligner matched the forced aligner, so alignment was not the cause. \textbf{E7} a Conformer-duration fine-tune produced junk through a blank-eating rescale. \textbf{E8} a pad-silence path was validated and then retired when the next family won.

\paragraph{Era 2, VITS2 (E9 to E18).}
A from-scratch end-to-end model was crisper than StyleTTS2 but kept the conjunct ceiling. \textbf{E9} adding the phoneme-level encoder was killed (the baseline was better on every metric). \textbf{E10} an end-to-end vocoder swap was killed (muffled). \textbf{E15} a second phoneme-level encoder was stopped for lack of a consumer. \textbf{E16} checkpoint averaging, plus the finding that a long-output explosion is seed-intrinsic, so a median-of-N seeds is needed. \textbf{E17} conjuncts stayed muffled in both models, traced to data sparsity for specific retroflex consonants. \textbf{E18} distributed-training validation.

\paragraph{Era 3, Matcha-TTS and vocoder (E19 to E28).}
\textbf{E19} a flow-matching acoustic model (18.2M). \textbf{E20} a vocoder bake-off: the large-scale adversarial vocoder won, and the Fourier-domain alternative was robotic. \textbf{E21} the pair reached MOS about 4.2 to 4.3 with seven residual issues, the most important being missing meter prosody. \textbf{E22} phase-two vocoder co-training on predicted mels. \textbf{E23} a multi-resolution-STFT vocoder won, and the repha was confirmed acoustic and therefore vocoder-fixable. \textbf{E24} a reframe toward indistinguishability and a plan for a 25 hour chant corpus. \textbf{E25} a corpus audit of two source texts. \textbf{E26} a rule-based meter-duration prior, to be baked as training conditioning rather than applied at inference. \textbf{E27} a recording manifest of 12{,}487 half-verse segments. \textbf{E28} a retroflex substitution test confirmed aspiration-specific data sparsity. A frozen Matcha configuration at about 4.2 to 4.3 was later superseded entirely.

\paragraph{Era 4, flow-matching backbone (E29 to E80h).}
\textbf{E29} zero-shot Sanskrit cloning worked once a normalization bug in the installed backbone was patched, and Devanagari schwa deletion was fixed by Kannada routing. \textbf{E30} a five hour fine-tune with a Kannada-versus-transliteration ablation. \textbf{E31} a long-verse split and the adoption of a 0.90 speed factor. \textbf{E32} the Kannada fine-tune reached about 4.6 MOS with all conjuncts, including retroflex aspirates, correct, and Kannada beat transliteration. \textbf{E33} a pivot to meter conditioning, with the 25 hour corpus deprioritized once it was clear that data was not the bottleneck. \textbf{E34} three residual tells, namely repeated-token collapse and a final-visarga drop. \textbf{E35} reference-prosody sensitivity confirmed, with a 1.6 times duration swing from the reference alone. \textbf{E36} an explicit meter-conditioning architecture was designed, namely a duration predictor and a length regulator. \textbf{E37} wobble was localized to inference, and the meter labeler was validated. \textbf{E38} to \textbf{E40} the meter featurizer and its backbone integration. \textbf{E41} implicit meter conditioning was inert (an on-versus-off ablation of plus 0.00 percent), which motivated the explicit route. \textbf{E42} forced alignment yielded ground-truth durations. \textbf{E43} a standalone duration predictor improved validation log-RMSE by about 26 percent. \textbf{E44} a vocoder fine-tune for the backbone on a 100-band mel. \textbf{E45} a vocoder reconstruction fine-tune on generated mels. \textbf{E46} a head and tail gate fixed the vocoder onset transient. \textbf{E47} the production-voice corpus was tallied at about 3.0 hours over 764 verses. \textbf{E48} the training-text normalizer, which is the released frontend. \textbf{E49} the production fine-tune was launched. \textbf{E50} a data-fix saga (a recording lead-in offset and a snap-to-gap refinement) and a clean relaunch. \textbf{E51} the reference was established as the voice and gross-meter path. \textbf{E52} a chant-paced duration predictor. \textbf{E53} the base-voice pick, the vocoder ruled out as the wobble source, and confirmation that within-verse pitch excursion is real at about 3.0 to 3.6 semitones. \textbf{E54} a swara conditioner appeared to work at high gain. \textbf{E55} generalization, tremor, vocal fry, and custom-swara transfer. \textbf{E56} a unified continuous conditioner. \textbf{E57} any-style paired data, where the swara overrode a conflicting reference, and a definitive test localized the wobble to the generator rather than the vocoder. \textbf{E58} an Arm A (plain) versus Arm B (normalized) comparison, and the rule never to split per quarter-verse because it severs cross-quarter sandhi. \textbf{E59} the vocoder fix, the conditioner dropped as inert, and the reference rules. \textbf{E60} deployment sizing and serving. \textbf{E61} humanization through a real breath at the da\d{n}\d{d}a and cropping of non-da\d{n}\d{d}a pauses. \textbf{E62} the gold champion snapshot, and a vocoder-class loading bug that silently half-loads the weights. \textbf{E63} the vocoder ceiling, with the gold soup locked. \textbf{E64} the serving configuration locked, and the finding that melody is seed-stochastic, so best-of-N is used. \textbf{E65} reference selection set to exact per-quarter-verse weight match, with the conditioner buried for want of a gradient. \textbf{E66} tempo transferred from the reference at about one to one, with more quarters in the reference giving more stability. \textbf{E67} classifier-free guidance identified as the tremor lever and raised to 3.0, with a dramatic reference as the swara lever. \textbf{E68} the swara embedding shown to be a definitive negative, because self-infilling reveals pitch from the context mel and the token is therefore redundant, never to be re-attempted. \textbf{E69} the visarga normalizer in the frontend. \textbf{E70} a half-reference repeated three times, to about 20 seconds, which markedly improves the swara. \textbf{E71} a semantic-token zero-shot synthesizer probed and paused for overfitting on a small corpus. \textbf{E72} to \textbf{E75} a description-conditioned model probed, with steerable prosody but content drift, then dropped. \textbf{E76} a long-vowel shiver identified as a vocoder phase artifact, with production locked on the adversarial vocoder and never the Fourier-domain one. \textbf{E77} the prosody wall, namely that the reference is a weak prosody carrier and the contour is text-driven. \textbf{E78} the text-side conditioner shown to train but not control, a proven dead end. \textbf{E79} the working lever, namely a voice-steering retrain plus the half-reference rule, which shipped. \textbf{E80} to \textbf{E80h} in-context priming for repeated syllables, a two-voice self-double, a tonic drone, gap and visarga rules, and internal sandhi turned on.

\section{Glossary}
v\d{r}tta: meter. gana: laghu and guru pattern. laghu and guru: light and heavy syllable. swara: melodic note or contour. da\d{n}\d{d}a: verse punctuation, half and full. yati: caesura. p\=ada: quarter-verse. hemistich: half-verse. sa\d{m}hit\=a: sandhified continuous text. sandhi: euphonic joining. visarga and anusv\=ara: two Sanskrit phonemes with orthographic marks. pram\=a\d{n}a: scriptural citation. MOS: mean opinion score. CFM: conditional flow matching.

\end{document}